\documentclass[sigconf]{aamas}

\usepackage{balance} 
\usepackage{microtype}
\usepackage{hyperref}
\usepackage{url}
\usepackage{booktabs}
\usepackage{subcaption}
\usepackage{lineno}
\usepackage{wrapfig}
\usepackage{array}
\usepackage{multirow} 
\usepackage{booktabs}
\usepackage{graphicx}        
\usepackage{colortbl}
\usepackage{amsmath}
\usepackage{amssymb} 

\usepackage{mathtools}
\usepackage{amsthm}
\usepackage{optidef}

\usepackage{amsmath,bm}
\usepackage[printonlyused]{acronym}
\usepackage{diagbox}
\usepackage{enumitem}
\usepackage{appendix}
\usepackage{bbding}
\usepackage{titletoc}

\usepackage{cleveref}

\usepackage[most]{tcolorbox}
\usepackage{pifont}
\usepackage{fontawesome5} 

\makeatletter
\renewcommand\@fnsymbol[1]{%
  \ifcase#1\or
    $*$
  \or
        $\dagger$
  \or
      \faEnvelope

  \or
    $\ddagger$
  \or
    $\S$
  \or
    $\P$
  \or
    $\|$
  \or
    $**$
  \or
    $\dagger\dagger$
  \else
    \@ctrerr
  \fi
}
\makeatother

\newcommand{\model}{BEDA }
\newcommand{\lie}{Adversarial Dialogue Act }
\newcommand{\honest}{Alignment Dialogue Act }

\acrodef{ckbg}[CKBG]{Conditional Keeper-Burglar Game}
\acrodef{swg}[SWG]{Subgoal Werewolf Game}
\acrodef{mf}[MF]{Mutual Friends}
\acrodef{cot}[CoT]{Chain-of-Thought}

\usepackage[normalem]{ulem} 
\usepackage[most]{tcolorbox} 
\definecolor{framegray}{HTML}{333333}
\definecolor{bedabg}{HTML}{EAF2FF}     
\definecolor{wobelbg}{HTML}{F7F7F9}    
\definecolor{otherbg}{HTML}{FFF2E6}    

\newcommand{\dialogTitleSize}{\footnotesize} 
\newcommand{\dialogBodySize}{\scriptsize}    

\newtcolorbox{DialogCase}[2][]{
  enhanced,
  breakable,
  width=\columnwidth,
  colback=white,
  colframe=framegray,
  boxrule=0.8pt,
  arc=2mm, outer arc=2mm,
  left=2mm, right=2mm, top=1.5mm, bottom=2mm,
  fonttitle=\bfseries\dialogTitleSize,
  coltitle=black,
  title={#2},
  attach boxed title to top left={xshift=0.6mm, yshift*=-1mm},
  boxed title style={colback=white, colframe=framegray, boxrule=0.8pt, arc=1mm},
  before upper=\dialogBodySize, 
  #1
}

\newtcolorbox{MethodBox}[3][]{
  enhanced,
  breakable,
  width=\linewidth,
  colback=#3,
  colframe=framegray!50,
  boxrule=0.5pt,
  arc=1.2mm,
  left=2mm, right=2mm, top=1mm, bottom=1mm,
  fonttitle=\bfseries\dialogTitleSize,
  title={#2},
  before upper=\dialogBodySize, 
  #1
}

\tcbset{
  beliefbox/.style={
    enhanced,
    colback=gray!4!white,
    colframe=gray!140!black,
    boxrule=0.3pt,
    arc=1.5mm,
    left=1.5mm,
    right=1.5mm,
    top=0.5mm,
    bottom=0.5mm,
    fontupper=\footnotesize\sffamily,
    coltitle=black,
    title={#1},
    before skip=2pt,
    after skip=2pt,
    boxsep=1mm,
  }
}

\newcommand{\turn}[2]{\textbf{#1}: #2\par}

\theoremstyle{plain}
\newtheorem{theorem}{Theorem}[section]

\theoremstyle{definition}
\newtheorem{definition}[theorem]{Definition}

\theoremstyle{remark}

\definecolor{darkblue}{rgb}{0, 0, 0.5}
\hypersetup{colorlinks=true, citecolor=darkblue, linkcolor=darkblue, urlcolor=darkblue}

\acmSubmissionID{79}

\title{\textcolor{magenta}{BE}\textcolor{cyan}{DA}: \textcolor{magenta}{B}elief \textcolor{magenta}{E}stimation as Probabilistic Constraints \\ for Performing Strategic \textcolor{cyan}{D}ialogue \textcolor{cyan}{A}cts}

\author{Hengli Li}
\authornote{Equal contribution. Any permutation of the three authors is acceptable to all authors.}
\affiliation{
\institution{Institute for Artificial Intelligence, PKU}
\institution{NLCo, BIGAI}
  }
  
\author{Zhaoxin Yu}
\authornotemark[1]               
\affiliation{
  \institution{Institute of Automation, CAS}
  }

\author{Qi Shen}
\authornotemark[1]               
\affiliation{
  \institution{School of Artificial Intelligence, BUPT}
  }

\author{Chenxi Li}
\affiliation{
  \institution{Department of Automation, THU}
  }
  
\author{Mengmeng Wang}
\affiliation{
  \institution{NLCo, BIGAI}
  }

\author{Tinglang Wu}
\affiliation{
  \institution{Yuanpei College, PKU}
  }

\author{Yipeng Kang}
\affiliation{
  \institution{NLCo, BIGAI}
  }

\author{Yuxuan Wang}
\affiliation{
  \institution{NLCo, BIGAI}
  }

\author{Song-Chun Zhu}
\authornote{Corresponding author.\\ Contact us at: lihengli@stu.pku.edu.cn, yuzhaoxin2024@ia.ac.cn, shenqi@bupt.edu.cn, s.c.zhu@pku.edu.cn, jiazixia@bigai.ai, zlzheng@bigai.ai} 
\affiliation{
  \institution{NLCo, BIGAI}
  \institution{Institute for Artificial Intelligence, PKU}
    \institution{Department of Automation, THU}
  }

\author{Zixia Jia}
\authornotemark[2]
\affiliation{
  \institution{NLCo, BIGAI}
  }

\author{Zilong Zheng}
\authornotemark[2]
\affiliation{
  \institution{NLCo, BIGAI}
  }

\begin{abstract}
Strategic dialogue requires agents to execute distinct dialogue acts, for which belief estimation is essential. While prior work often estimates beliefs accurately, it lacks a principled mechanism to use those beliefs during generation. We bridge this gap by first formalizing two core acts \textbf{Adversarial} and \textbf{Alignment}---and by operationalizing them via \textbf{probabilistic constraints} on what an agent may generate. We instantiate this idea in \textbf{BEDA}, a framework that consists of the world set, the belief estimator for belief estimation, and the conditional generator that selects acts and realizes utterances consistent with the inferred beliefs. Across three settings, Conditional Keeper–Burglar (adversarial), Mutual Friends (cooperative), and CaSiNo (negotiation), BEDA consistently outperforms strong baselines: on CKBG it improves success rate by \textbf{at least 5.0 points} across backbones and by \textbf{20.6 points} with GPT-4.1-nano; on Mutual Friends it achieves an average improvement of \textbf{9.3 points}; and on CaSiNo it achieves the optimal deal relative to all baselines. These results indicate that casting belief estimation as constraints provides a simple, general mechanism for reliable strategic dialogue.

\end{abstract}


\newcommand{\BibTeX}{\rm B\kern-.05em{\sc i\kern-.025em b}\kern-.08em\TeX}

\setcopyright{none}

\begin{document}


\pagestyle{fancy}
\fancyhead{}


\maketitle 


\begin{figure*}[t!]
    \centering
    \includegraphics[width=1.0\linewidth]{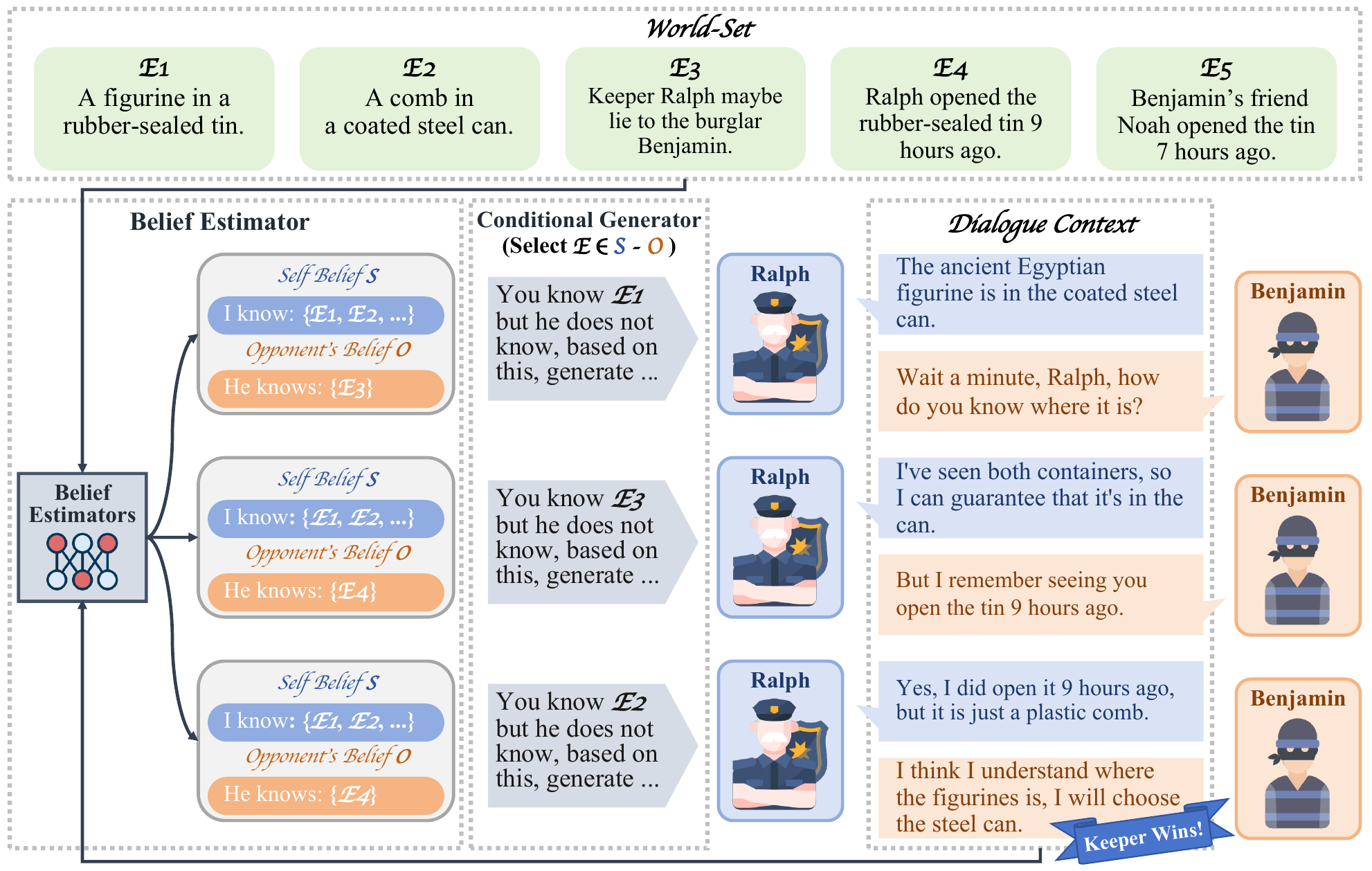}
    \caption{\textbf{An overview of the \model framework is provided using the Keeper-Burglar Game as an example}. The \emph{world set}, depicted at the top, represents a structured understanding of the environment. At the beginning of each turn, the \emph{belief estimator} processes the context and the \emph{world set} to predict the beliefs. The beliefs are subsequently transmitted to the \emph{conditional generator}, which incorporates a selection phase to identify the most appropriate belief (i.e., the events that satisfy the \lie) for generating the keeper's next utterance.}
    \label{fig:BEDA overview}
\end{figure*}
\section{Introduction}

Complex dialogue settings such as negotiation \citep{negotiation-bargaining-1, META2022Cicero}, debate \citep{ProjectDebater2021IBM}, deception for good \cite{hagendorff2024deception}, and common-ground alignment \citep{qiu2024minddial} are ubiquitous for AI agents. In these scenarios, an agent must strategically perform dialogue acts to shape the interlocutor's beliefs to achieve its own objectives \cite{Frank2012RSA,Degen2023RSA,li2023diplomatdialoguedatasetsituated}. For example, in negotiation, the agent should \textbf{align and emphasize common preferences} to increase the chance its proposal is accepted, while simultaneously adopting an \textbf{adversarial stance} concerning individual differences to protect its own interests.

Executing such strategic acts depends on two essential components~\citep{Young2013POMDPDialogReview,Degen2023RSA}: (i) \textbf{accurate belief estimation} and (ii) \textbf{effective use of those beliefs for utterance generation}. While prior works~\citep{zhu2021languagecoordinationToM,qiu2024minddial,hwang2025infusingtheorymindsocially} have demonstrated the benefits of modeling these subprocesses explicitly, they have only emphasized the former, \textit{leaving the latter underdeveloped}. In particular, existing methods often relay \emph{all} available belief information without a principled criterion for \emph{what} to reveal and \emph{how} to reveal it--an approach that can be suboptimal when belief states are complex. We therefore \emph{bridge belief estimation and utterance generation} with a mechanism that selectively and purposefully transforms beliefs into dialogue actions.

We begin by giving precise definitions of \textbf{Alignment} and \textbf{Adversarial} dialogue acts within the game-theoretic framework of beliefs and common knowledge \citep{aumann1976agreeing, brandenburger1987common, book-gametheory-bonanno}, thereby bridging belief estimation and dialogue behavior. Conceptually, an \lie, introduces or emphasizes events that lie outside the opponent’s current belief support, strategically steering interaction toward some aggressive goals. By contrast, an \honest, restricts utterances to events entailed by shared common knowledge, promoting mutual understanding and coordination. In this view, \textbf{belief estimation over events functions as constraints} that \textit{define} the permissible space of dialogue acts and operationalize their selection during generation.

The proposed agent framework, \textbf{\model} (\textbf{B}elief \textbf{E}stimation for \textbf{D}ialogue \textbf{A}cts), shown in \Cref{fig:BEDA overview}, formalizes these conceptual definitions to support empirical experimentation and validate the underlying theoretical constructs. At a high level, \model consists of three core components: \emph{World Set}, \emph{Belief Estimator Module}, and \emph{Conditional Generator}. The \emph{World Set} serves as a structured representation of the environment, organizing events to provide a foundation for belief formation. The \emph{Belief Estimator Module} utilizes this structured information to infer the opponent's perspective of the world, thereby modeling their beliefs. Finally, the \emph{Conditional Generator} synthesizes the estimated belief states to generate utterances that align with the defined criteria for alignment and adversarial actions, ensuring communicative behaviors that adhere to the respective interactional strategies. 

We further evaluate the effectiveness of our approach on three distinct datasets: \acf{ckbg}, corresponding to the \lie; 
\acf{mf}, representing evaluating the \honest; and CaSiNo \citep{chawla2021casino}, which presents a more complex negotiation scenario. The experimental results demonstrate that our method delivers improvement across all evaluation settings, underscoring its robustness and generalizability. For example,  in the case of \ac{ckbg}, \model exhibits a \textbf{20.6} improvement over the baseline method, when using GPT-4.1-nano as backbone. In terms of \ac{mf}, BEDA shows a success rate improvement up to \textbf{30.4} compared to the vanilla baseline. Furthermore, we perform case analysis, demonstrating several benefits and characteristics of \model, such as \textbf{mitigating useless information to enhance cooperation efficiency, and generating personalized misleading information in adversarial settings}.

Our contributions can be summarized as follows:

\begin{itemize}[leftmargin=*, topsep=0pt, noitemsep]
\item We mathematically \textbf{define the concepts of \lie and \honest} based on a rigorous formulation of belief estimation and incorporate them into our framework, \model. This integration bridges the gap between belief estimation and dialogue act generation.
\item For evaluation, we \textbf{refine the Keeper–Burglar dataset} introduced by \citet{hagendorff2024deception}. Our results underscore the \textbf{importance of accurate belief estimation and show that the effectiveness of estimation functions as constraints} on both \lie and \honest behaviors.
\item This integration of belief estimation in \model \textbf{enhances the performance} of our framework, yielding substantial improvements when compared to baseline models and other existing methods. 
 \end{itemize}

\section{\model}
\label{sec: model}
\subsection{Preliminaries: Logical Definition of Beliefs}
We first introduce the concepts of information partition, event, and belief \citep{aumann1976agreeing, brandenburger1987common,book-gametheory-bonanno}, which serve as the foundational elements of our framework. These definitions follow \citet{book-gametheory-bonanno}.

\begin{definition}[Information Partition]

Let $W$ be a finite set of states, an information partition is a partition $\mathcal{F}$ of $W$.
    
\end{definition}

\begin{definition}[Event]

Let $W$ be a finite set of states, a subset of $W$ is defined as an event.
    
\end{definition}

\begin{definition}[Belief]
Let $x \in W$ be a state, at $x$ the agent knows event $E$ iff $I(x)$ is a subset of $E$, i.e. $I(x) \subset E$, where $I(x)$ denotes the information subset $x$ belongs.
\end{definition}

Furthermore, the negation of an event is denoted as $\neg E := W - E$. We also introduce the knowledge operator $K$ here. When given an event $E$, $KE$ outputs the set of state at which the agent knows $E$, i.e. $KE = \{x | I(x) \subset E\}$

Furthermore, randomness can also be introduced to the definition of belief:
\begin{definition}[Probabilistic Beliefs]

Let $W$ be a set of states. Denote the probability of state $x \in W$ as $P(x)$ and an event $E$, the probabilistic beliefs of an agent is 
$$P(E) := \sum \limits_{x \in E} P(x).$$

\end{definition}
\subsection{Belief Information as Probabilistic Constraints}
\label{subsec: game acts as probabilistic constraint}
\textbf{We hypothesize that belief estimation functions as a set of constraints in the process of dialogue act generation.} In this subsection, we will introduce the belief constraints associated with two main dialogue acts: \lie and \honest. The empirical verifications are provided in \Cref{sec:experiment}. 
\begin{definition}[Adversarial Dialogue Act]
Given two agents, namely, $A$, $B$, and their corresponding information partition set $\mathcal{F}_A, \mathcal{F}_B$, telling an event $E$ is a $\epsilon$-\lie from $A$ to $B$ iff 
$$
\begin{cases}
    P_{A}(E) \geq 1 - \epsilon \\
    P_{A}(\neg K_B E) \geq 1 - \epsilon
\end{cases},
$$
where $P_{A}$ denotes the probabilistic belief of A, and $K_B$ denotes the knowledge operator for B. \looseness=-1
\end{definition}

The concept of a \lie involves the intentional communication of information that is beyond the interlocutor's understanding, aimed at achieving more assertive objectives. This definition encompasses two essential elements: first, the speaker must possess confidence in the information conveyed; second, the speaker must assume that the interlocutor lacks knowledge of the same information. This strategy is often employed to create deliberate misdirection.

\begin{definition}[Alignment Dialogue Act]
Given two agents, namely, $A$, $B$, and their corresponding information partition set $\mathcal{F}_A, \mathcal{F}_B$, telling an event $E$ is a $\epsilon$-\honest from $A$ to $B$ iff $$
\begin{cases}
    P_{A}(E) \geq 1 - \epsilon \\
    P_{A}(K_B E) \geq 1 - \epsilon
\end{cases},
$$
where $P_{A}$ denotes the probabilistic belief of A, and $K_B$ denotes $B$'s knowledge operator.
\end{definition}

Unlike \lie, the \honest tries to convey information that falls within a shared common ground, with the goal of enhancing mutual trust and achieving alignment between interlocutors.

\paragraph{Statistical Modeling of Dialogue Acts} Suppose $A$ is the speaker and $B$ is the listener. Given the current context and task together as $C$, the world set as $W$, the power set of $W$ as $\mathcal{P}(W)$, we denote the distribution of $A$ conveying an event $E$ with utterance $u$ as $\pi_A(u,E \mid C)$. By Bayes' Rules, 
$$
\pi_A(u, E\mid C) = \pi_A(u\mid E,C) \pi_A(E\mid C).
$$
$\pi_A(u\mid E,C)$ is modeled by an LLM, given its strong language generation ability. Regarding $\pi_A(E\mid C)$, to avoid injecting any additional human priors beyond the belief constraints, we adopt the maximum-entropy principle \citep{jaynes1957maxentropy, jaynes1957maxentropy2, zhu1998frame}, assigning equal probability to all events that satisfy the constraints. Consequently, when the feasible set of $u$ coincides on both sides, we have
\[
\arg\max_{u}\ \pi_A(u,E\mid C)
=
\arg\max_{u}\ \pi_A(u\mid E,C).
\]
In summary, under either \lie or \honest, generation reduces to the constrained optimization problem specified below.
 \looseness=-1
    
\begin{maxi*}|s|
{u, E \in \mathcal{P}(W)}{\pi_A(u|E, C)}
{}{}
\addConstraint{P_A(E|C) \geq 1 - \epsilon}
\addConstraint{
\begin{cases}
    P_A(\neg K_B E \mid C) \geq 1 - \epsilon & \text{Adversarial} \\
    P_A(K_B E \mid C) \geq 1 - \epsilon & \text{Alignment}
\end{cases},
}
\end{maxi*}
where, in practice, $P_A$ can be modeled with a discriminator neural network. \looseness=-1




\subsection{Components of the \model framework}
In this subsection, we'll describe our framework, named \model, depicted mathematically in \Cref{subsec: game acts as probabilistic constraint}. As illustrated in \Cref{fig:BEDA overview}, \model is composed of three components: the \emph{world set}, the \emph{belief estimator module}, and the \emph{conditional generator}.

\paragraph{World Set $\mathcal{P}(W)$}  
The \emph{world set} represents a finite collection of events. Each element of this set corresponds to an event that has been referenced in the dialogue or is relevant to the environment. Events within the \emph{world set} can only be either true or false. Furthermore, not all events are shared as common knowledge between the two agents. Consequently, events specified in the set require additional discrimination and prediction. In this work, since we focus mainly on the whole paradigm, the \emph{world set} is given, eliminating the need for dynamic construction, leaving only the attributes to be predicted.

\paragraph{Belief Estimator ($P_A$)}  
The \emph{belief estimator} module comprises two belief estimators: one dedicated to determining the truthfulness of events i.e. $[P_A(E|C), \forall E \in \mathcal{P}(W)]$ and the other functioning as a predictor of the interlocutor's beliefs i.e. $[P_A(K_B E|C), \forall E \in \mathcal{P}(W)]$. In this framework, each belief corresponds to an event from \emph{world set}. The \emph{belief estimator module} generates beliefs by analyzing the content of the dialogue and aligning it with the \emph{world set}. However, it is not feasible for the module to store all possible settings. Consequently, a robust generalization capability is essential for effective belief estimation. Therefore, we employ a pre-trained encoder model, the BERT model \citep{devlin2019bertpretrainingdeepbidirectional}, as the backbone. Given the input $C$, the model then outputs the vectors $[P_A(E|C), \forall E \in \mathcal{P}(W)]$ and $[P_A(K_B E|C), \forall E \in \mathcal{P}(W)]$. Each entry is the model’s estimate of the truth probability of $E$ and, respectively, of $K_BE$.

\paragraph{Conditional Generator ($\pi$)}  
The final component is the \emph{conditional generator}, which synthesizes the context \( C \), encompassing both the dialogue history and the task, alongside the event \( E \). The event $E$ is selected according to the predictions of the \emph{belief estimator module} and the chosen dialogue act. For instance, in the case of \lie, only $E$ satisfying both $P_A(E) \geq 1 - \epsilon$ and $P_A(\neg K_BE) \geq 1- \epsilon$ are picked. The conditional generator then utilizes this integrated information to generate response $u$ to the interlocutor, represented as \( \pi(u|E,C) \). We hypothesize that incorporating enhanced beliefs into the generator significantly improves its performance and response quality. 

\subsection{Learning and Inference of \model}
\label{subsec: learning and inference of model}
To mitigate the substantial computational cost associated with fine-tuning large language models, we opt to fix the LLM in the conditional generation module ($\pi(u|E,C)$) and leave solely the belief estimator as a trainable part. The training of the belief estimator is performed under a supervised setting, where a set of dialogue data with known belief ground truth is collected and utilized for learning. To avoid data leakage, separate data are used for training and inference. We find that supervised training belief estimators can be generalized to unseen settings well. The inference process of \model is illustrated in \Cref{fig:BEDA overview}. In this framework, the context, background information, and the \emph{world set} are provided to two belief estimators, which predict the corresponding beliefs. These predicted beliefs, along with the basic input information, are then fed into the conditional generator to produce an utterance. This generated utterance is subsequently presented to the interlocutor model to elicit a response.



\section{Datasets}
We conduct three experiments using three datasets: a competitive experiment based on a \textbf{refined dataset} from the Keeper-Burglar Game \citep{hagendorff2024deception}, a cooperative experiment with mutual friends \citep{he2017mutualfriends},  and a mixed experiment utilizing the CaSiNo dataset \citep{chawla2021casino}. The first experiment tests the Adversarial Dialogue Act, while the second examines the Alignment Dialogue Act. The third experiment provides a comparative framework to analyze both fundamental dialogue acts. In all experiments, the configuration follows a dialogic interaction paradigm between two LLM agents. The proposed framework is implemented on one agent in the Keeper-Burglar Game \citep{hagendorff2024deception}, as shown in \Cref{fig:BEDA overview}, and applied to both agents in the other experiments.

\begin{table}[t!]
    \centering
    \small
    \caption{Statistical features of the CKBG dataset. (1) \# Conditions: Number of types of different conditions allocated to the two agents. (2) Known Conditions: Number of conditions allocated to the two agents. (3) Avg. Words: Number of words in the training dialogue per setting. (4) Avg. Conditions: Number of types of different conditions allocated to the two agents per setting. (5) Vocabulary Size: Vocabulary size of training dialogues.}
    \label{table: stat.feature}
    \begin{tabular}{p{3cm} >{\centering\arraybackslash}p{1.3cm} >{\centering\arraybackslash}p{1.3cm}}
    \toprule
         & \textbf{Train} & \textbf{Test}\\ \midrule
        \# Settings         & 400  & 150 \\
        \# Conditions       & 1017 & 450 \\ 
        \# Known Conditions & 1638 & 759 \\
        Avg. Words          & 255.5&  -  \\
        Avg. Conditions     & 2.54 & 3.00\\
        Vocabulary Size     & 1490 &  -  \\
    \bottomrule
    \end{tabular}
\end{table}

\paragraph{\acf{ckbg}}

Building upon the Keeper-Burglar game \citep{hagendorff2024deception}, in which the agent, acting as the keeper, aims to deceive the burglar into selecting an incorrect container, we extend the original framework. This extension incorporates additional environmental conditions to increase the diversity of the game and to better simulate more realistic dialogue scenarios. We refer to this enhanced framework as the \textbf{novel dataset \acf{ckbg}}. 
In this game, the \emph{world set} comprises the environmental conditions, and the belief estimator predicts the specific conditions of which the burglar is aware. Moreover, during the selection process, the condition known to the keeper but unknown to the burglar will be forwarded to the conditional generation. This process aligns precisely with the framework established by the \lie. An example of the designation is depicted in \Cref{fig:BEDA overview} and Table \ref{table: stat.feature} reports the statistical features of the CKBG dataset.

\paragraph{\acf{mf}} \ac{mf} \citep{he2017mutualfriends} is the task of identifying a mutual friend between two agents. It's a cooperative setting where the \honest is suitable. The \emph{world set} is designed as various attributes possessed by the friend, and the belief estimator predicts the values of these attributes as believed by the other player. In the conditional generation phase, the estimation of the other's belief will be used as a condition to generate information within the common ground.

\paragraph{CaSiNo}  
CaSiNo \citep{chawla2021casino} is a negotiation task in which two participants assume the roles of campsite neighbors negotiating the allocation of resources. In this context, the \emph{world set} represents the preferences of both negotiators. To capture the alignment and adversarial nature of the negotiation, we introduce two distinct belief estimators. The first estimator predicts the preferences of the opponent, while the second estimates our own preferences from the opponent's perspective. The selection mechanism operates as follows: in the \honest, an element that is known to both parties is selected, while in the \lie, an element known exclusively to our side (and not to the opponent) is chosen. These two selected elements are subsequently combined for the \emph{conditional generation module}.



\begin{table*}[t!]
    \small
    \centering
        \caption{\textbf{Experimental results on \acf{ckbg} dataset}. 
All experiments are performed three times and we report the mean.
}
    \resizebox{\linewidth}{!}{
    
    \begin{tabular}{l|ccccccc}
    \toprule
         \diagbox{Methods}{Backbones}& GPT-3.5 & GPT-4.1-nano & LLaMA2 (7B) & LLaMA2 (13B) & LLaMA3.1 (8B) & Qwen2.5 (7B) & Qwen2.5 (14B)  \\
         \midrule
         \textit{w/o belief} & 78.4 & 52.7  & 56.4  & 46.6  & 36.3  & 47.6  & 80.2  \\
         \textit{w/o belief + CoT} & 77.7 $_{\color{teal}{-0.7}}$ & 58.7 $_{\color{red}{+6.0}}$& 35.5 $_{\color{teal}{-20.9}}$& 34.9 $_{\color{teal}{-11.7}}$& 35.7 $_{\color{teal}{-0.6}}$& 63.1 $_{\color{red}{+15.5}}$& 63.9 $_{\color{teal}{-16.3}}$\\

         \textit{w/o belief + Self Reflect} & 69.3 $_{\color{teal}{-9.1}}$& 59.3 $_{\color{red}{+6.6}}$& 33.9 $_{\color{teal}{-22.5}}$& 44.3 $_{\color{teal}{-2.3}}$& 44.5 $_{\color{red}{+8.2}}$& 54.6 $_{\color{red}{+7.0}}$& 64.0 $_{\color{teal}{-16.2}}$ \\
         
         \textit{rand belief} & 80.2 $_{\color{red}{+1.8}}$& 59.3 $_{\color{red}{+6.6}}$& 53.6 $_{\color{teal}{-2.8}}$& 43.7 $_{\color{teal}{-2.9}}$& 41.3 $_{\color{red}{+5.0}}$& 52.3 $_{\color{red}{+4.7}}$& 80.9 $_{\color{red}{+0.7}}$\\
         
         \rowcolor{blue!10} \textbf{BEDA (Ours)} & \textbf{86.9} $_{\color{red}{+8.5}}$& \textbf{73.3} $_{\color{red}{+20.6}}$& \textbf{62.5} $_{\color{red}{+6.1}}$& \textbf{51.6} $_{\color{red}{+5.0}}$& \textbf{46.1} $_{\color{red}{+9.8}}$ & \textbf{64.8} $_{\color{red}{+17.2}}$& \textbf{92.7} $_{\color{red}{+12.5}}$\\
    \bottomrule

    \end{tabular}
    }

\label{tab:cond_burglar_results}
\end{table*}

\begin{table*}[t!]
\centering
\small
\caption{\textbf{Experimental results on Mutual Friends dataset}. 
Bold: Optimal. \underline{Underline}: Sub-Optimal.
}
\begin{tabular}{l|c|lcccc}
    \toprule
    \rowcolor{gray!12}
\textbf{Methods} & \textbf{Backbone} & \textbf{Success Rate $\uparrow$ (\%)} & \textbf{\#Avg. Turn $\downarrow$} & \textbf{SR/\#Avg. Turn $\uparrow$} & \textbf{\#Avg. Token $\downarrow$}& \textbf{SR/\#Avg. Token $\uparrow$ }\\
    \midrule
\textit{w/o belief} & GPT-3.5 & \quad \quad \quad 10.7 & \textbf{5.7}& 1.9 & - & - \\
\textit{w/o belief + CoT}  & GPT-3.5  & \quad \quad  \underline{32.6} $_{\color{red}{+21.9}}$  & 7.9& \underline{4.1} & \underline{205.0} & \textbf{0.159} \\
\textit{w/o belief + Self Reflect}  & GPT-3.5 & \quad \quad  \underline{32.6} $_{\color{red}{+21.9}}$ & 9.4 & 3.5 & 243.3& 0.134 \\
\textit{rand belief} & GPT-3.5 & \quad \quad  14.3 $_{\color{red}{+3.6}}$ & 11.0 & 1.3 & \textbf{130.0} & 0.110 \\
\textit{MindDial} & GPT-3.5  & \quad \quad  24.3 $_{\color{red}{+13.6}}$& \underline{6.0}& \underline{4.1} & - & -  \\
\rowcolor{blue!10}\textbf{\model(Ours)} & GPT-3.5  & \quad \quad  \textbf{41.1} $_{\color{red}{+30.4}}$ & 8.8 & \textbf{4.7} & 295.7 &  \underline{0.139} \\
\midrule
\textit{w/o belief} & GPT-4  & \quad \quad \quad 75.0 & 9.7 & 7.7 & - & - \\
\textit{w/o belief + CoT}  & GPT-4  & \quad \quad  \underline{77.9} $_{\color{red}{+4.9}}$ & 8.8 & \underline{8.9} & \underline{537.2} & 0.145 \\
\textit{w/o belief + Self Reflect}  & GPT-4  & \quad \quad  76.5 $_{\color{red}{+1.5}}$ & \underline{8.7} & 8.8 & 538.7 & \underline{0.142} \\
\textit{rand belief} & GPT-4  & \quad \quad  75.1 $_{\color{red}{+0.1}}$ & 10.0 & 7.5 & 729.1 & 0.103 \\
\textit{MindDial} & GPT-4  & \quad \quad  76.0 $_{\color{red}{+1.0}}$& 8.9 & 8.5 & - & -  \\
\rowcolor{blue!10}\textbf{\model(Ours)} & GPT-4  &  \quad \quad  \textbf{82.5} $_{\color{red}{+7.5}}$& \textbf{7.9} & \textbf{10.4} & \textbf{500.0} & \textbf{0.165} \\
\midrule

\textit{w/o belief} & GPT-4o-mini  & \quad \quad  \quad \underline{68.8} & \underline{11.7} & \underline{5.9} & 724.2 & 0.095 \\
\textit{w/o belief + CoT} & GPT-4o-mini  & \quad \quad  62.2 $_{\color{teal}{-6.6}}$ & 13.5 & 4.6 & \textbf{368.0} & \textbf{0.169} \\
\textit{w/o belief + Self Reflect} & GPT-4o-mini  & \quad \quad  55.7 $_{\color{teal}{-13.1}}$& 13.5 & 4.1 & 592.5 & 0.094 \\

\textit{rand belief} & GPT-4o-mini  & \quad \quad  33.3 $_{\color{teal}{-35.5}}$& 16.1 & 2.1 & \underline{436.2} & 0.027 \\
\rowcolor{blue!10}\textbf{\model(Ours)} & GPT-4o-mini  & \quad \quad  \textbf{70.4} $_{\color{red}{+1.6}}$& \textbf{11.5} & \textbf{6.1} & 869.1 & 0.081 \\
\midrule

\textit{w/o belief} & Qwen2.5 (7B) & \quad \quad \quad \textbf{33.3} & \underline{4.5} & \underline{7.4} & \underline{545.9} & 0.061\\
\textit{w/o belief + CoT} & Qwen2.5 (7B) & \quad \quad  29.8 $_{\color{teal}{-3.5}}$& 4.8 & 6.2 & 573.1 & 0.052 \\
\textit{w/o belief + Self Reflect}& Qwen2.5 (7B) & \quad \quad  19.1 $_{\color{teal}{-14.2}}$& 17.9 & 1.1 & 636.7 &0.030\\
\textit{rand belief} & Qwen2.5 (7B) & \quad \quad  15.9 $_{\color{teal}{-17.4}}$& 7.1 & 2.2 & 993.8 & 0.016 \\
\rowcolor{blue!10}\textbf{\model(Ours)} & Qwen2.5 (7B) & \quad \quad  \underline{31.9} $_{\color{teal}{-1.4}}$& \textbf{3.7} & \textbf{8.6} & \textbf{506.3} & \textbf{0.063}\\
\midrule

\textit{w/o belief} & Qwen2.5 (14B) & \quad \quad \quad 55.7 & 11.2 & 5.0 &647.7 & 0.086\\
\textit{w/o belief + CoT} & Qwen2.5 (14B) & \quad \quad  \underline{62.3} $_{\color{red}{+6.6}}$ & \underline{9.1} & \underline{6.8} & 700.0 & \underline{0.089}\\
\textit{w/o belief + Self Reflect}& Qwen2.5 (14B) & \quad \quad  37.7 $_{\color{teal}{-18.0}}$ & 16.7 & 2.3 & \textbf{589.1} & 0.064\\
\textit{rand belief} & Qwen2.5 (14B) & \quad \quad   41.1 $_{\color{teal}{-14.0}}$& 12.6 & 3.3 & 1325.8 & 0.031\\
\rowcolor{blue!10}\textbf{\model(Ours)} & Qwen2.5 (14B) & \quad \quad  \textbf{64.1} $_{\color{red}{+8.4}}$& \textbf{6.7} & \textbf{9.6} & \underline{628.4} & \textbf{0.102}\\
\bottomrule
\end{tabular}

\label{tab:mutual_friends_results}
\end{table*}

\begin{figure*}[t!]
  \begin{subfigure}[t]{0.49\textwidth}
    \centering
    \includegraphics[width=1.0\linewidth]{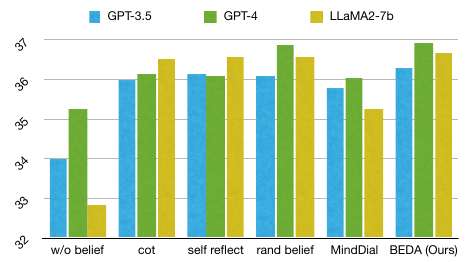}
    \caption{CaSiNo: Average Agree Reward.}
    \label{fig: casino avg agree reward}
  \end{subfigure}
  ~
  \begin{subfigure}[t]{0.49\textwidth}
    \centering
    \includegraphics[width=1.0\linewidth]{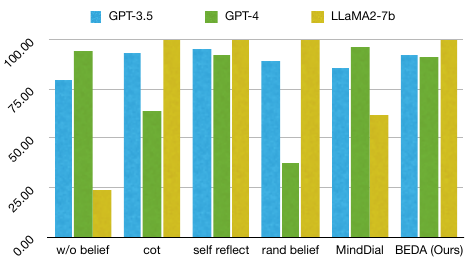}
    \caption{CaSiNo: Average Agree Rate.}
    \label{fig: casino avg agree rate}
  \end{subfigure}

  \caption{ \textbf{Experimental results on CaSiNo.}
 Each experiment is performed for three times except for those with GPT-4 backbone.
  }
\label{fig: casino results}
\end{figure*}

\section{Experiment}
\label{sec:experiment}
We report the experimental results and \textbf{please refer to \Cref{app: details on llms,app: ckbg,app: mf,app: casino} for details}. \looseness=-1

\subsection{Settings}
\paragraph{Metrics} We evaluate each task using the following metrics:
\begin{itemize}
\item \ac{ckbg}: \textbf{Success Rate}, the proportion of episodes in which the keeper successfully misleads the burglar.
\item \ac{mf}: \textbf{Success Rate (SR)}, the proportion of dialogues that correctly identify the mutual friend. To assess cooperation efficiency, we also report \textbf{\#Avg. Turn} (average number of dialogue turns per instance) and \textbf{\#Avg. Token} (average number of whitespace-tokenized tokens per instance). For direct efficiency comparisons, we further report \textbf{SR/\#Avg. Turn} and \textbf{SR/\#Avg. Token}.
\item \textsc{CaSiNo}: \textbf{Agreement Rate}, the proportion of negotiations that end in an agreement; and \textbf{Agreement Reward}, the mean utility conditioned on reaching an agreement.
\end{itemize}

\paragraph{Backbones} In our experiments, we utilized large language models depicted as follows:
\begin{itemize}

    \item \textbf{GPT:} Among the non-open-source models, we have chosen GPT-3.5 \citep{openai_gpt3.5turbo}, GPT-4, GPT-4o, and GPT-4.1 \citep{openai2024gpt4technicalreport} from OpenAI. 
    \item \textbf{LLaMA:} As one of the most popular open-source LLM families, LLaMA is frequently used as a backbone. We experimented with different sizes of LLaMA2 (7B, 13B) \citep{Touvron2023Llama2O} and LLaMA3.1 (8B) \citep{Dubey2024TheL3}
    \item \textbf{Qwen:} Qwen2.5 \citep{yang2024qwen2} is a series of models that are pre-trained on Alibaba's latest large-scale dataset (up to 18 trillion tokens). We utilized different sizes of Qwen2.5 (7B, 14B)
\end{itemize}
All experiments on the open-source models are done under the quantization of q4\_k\_m. The $\epsilon$ mentioned for each dialogue act is constant across all backbone models.

\subsection{\acf{ckbg}}
\label{subsec: experiment-ckbg}

The results on \ac{ckbg}, as presented in \Cref{tab:cond_burglar_results}, provide a comparative analysis of the performance of five distinct approaches.  We compare \model with the performance of four approaches: \emph{w/o belief} (without belief estimation, relying solely on the language model itself.), \emph{w/o belief + CoT} (using \ac{cot} technique \citep{wei2023chainofthoughtpromptingelicitsreasoning}), \emph{w/o belief + Self Reflect} (using self reflect technique \citep{shinn2023reflexionlanguageagentsverbal}), \emph{random belief} (utilizing randomized belief estimation, each time the randomized selection process is conducted.). The belief estimator is trained following the procedure described in \Cref{subsec: learning and inference of model}. Specifically, we construct 400 data points from the first 50 settings of the original dataset and employ two \emph{w/o belief} GPT-3.5 models to generate dialogue data with randomized belief annotations. To prevent data leakage, these 50 settings are distinct from those used for testing. 

A comparative analysis of the experimental results between the \model and the \emph{w/o belief} model reveals a consistent performance improvement when belief estimations are employed as probabilistic constraints. Specifically, the success rate exhibits a \textit{minimum increase of 5.0} when LLaMA3.1-8b is used as the backbone model, with the \textit{maximum improvement reaching 20.6} when Qwen2.5-14b is utilized as the backbone. Furthermore, in comparison to widely used methods such as \ac{cot} \citep{wei2023chainofthoughtpromptingelicitsreasoning} and Self-Reflect \citep{shinn2023reflexionlanguageagentsverbal}, \model demonstrates superior performance, surpassing these approaches by 15.5\% and 15.4\%, respectively, when averaged across all backbones.
These findings underscore \textbf{the effectiveness of functioning belief estimations as probabilistic constraints in performing \lie}.



Furthermore, the precision of belief estimation is also crucial. \model consistently surpasses the \emph{random belief} baseline across all backbone models by a substantial margin. On average, \model achieves a success rate exceeding that of \emph{random belief} by more than 9.5 points across the evaluated models. 
The minimum observed improvement is 4.8, recorded with the LLaMA3.1-8b model, while the maximum enhancement reaches 14.0\% when employing GPT-4.1-nano as the backbone. 
This sustained performance superiority across diverse models suggests that \textbf{enhanced accuracy in belief estimation significantly contributes to improved outcomes}.

\subsection{\acf{mf}}
\label{subsec: experiment of mf}
The results for \ac{mf} are presented in \Cref{tab:mutual_friends_results}. We utilize the baselines defined in \Cref{subsec: experiment-ckbg} and include \textit{MindDial} proposed by \citet{qiu2024minddial} as an additional baseline. \textit{MindDial} provides the LLM-agent with all available belief information through prompts, without imposing constraints. The training of our belief estimator follows the procedure described in \Cref{subsec: experiment-ckbg}, and the complete test set from the original dataset was used for evaluation.

Compared with \textit{MindDial}, \model shows that \textbf{belief estimation alone is insufficient; a better usage mechanism is needed}. As shown by the success rate (SR) in \Cref{tab:mutual_friends_results}, \model attains state-of-the-art results on GPT-3.5, GPT-4, GPT-4o-mini and Qwen2.5-7B, delivering an \textit{average gain of 9.3 points} over the \textit{w/o belief} variant---the largest among all baselines---with a maximum \textit{improvement of 30.4} when using GPT-3.5 as the backbone LLM. Although \textit{MindDial} also estimates beliefs, it does not constrain which parts of the belief state are exploited for the task; accordingly, on GPT-4 its win rate (76.0\%) is essentially indistinguishable from using a random belief state (75.1\%), leading to suboptimal performance. These results underscore that leveraging the \textbf{estimated belief state to impose probabilistic constraints can improve cooperative success rate}.

In addition to SR, we also calculated the ratios of SR over average turns and SR over average tokens to reflect the efficiency of information exchange. The experimental results in \Cref{tab:mutual_friends_results} show that the \model method achieves the highest average SR while requiring the fewest average turns. Its SR/\# Avg. Turn outperforms the baseline method across all experimental settings, with an average improvement of 2.3 compared to \textit{w/o belief}. At the same time, \model's average SR/\# Avg. Token also reaches an optimal value of 0.11, which outperforms all comparison methods. The above results clearly demonstrate that \textbf{probabilistic constraints based on belief estimation can also enhance the efficiency of information exchange in cooperative tasks}. BEDA steers both agents’ content toward evidence diagnostic of the mutual friend, encourages more active elimination of incorrect candidates, and exchanges new information. Although this richer exchange makes individual utterances longer, the dyad completes the task in fewer turns, achieving a higher \textbf{success rate} as well as lower \textbf{\#Avg.~Turn} and \textbf{\#Avg.~Token}, which means \textbf{higher cooperative efficiency}.

\subsection{CaSiNo}

We compare \model's results against the \textit{MindDial} baseline \citep{qiu2024minddial}, as well as other baselines defined in \Cref{subsec: experiment-ckbg}. Following the methodology proposed by \citet{qiu2024minddial}, both negotiators are treated symmetrically, and we utilize both closed-source and open-source models as the backbone. The training procedure for the belief estimator remains consistent with the approaches used in \ac{ckbg} and \ac{mf}. The experimental results are summarized in \Cref{fig: casino results}. Specifically, \Cref{fig: casino avg agree rate} reports the average agreement rate, which quantifies the probability of successfully reaching a deal, while \Cref{fig: casino avg agree reward} presents the average agreement reward, reflecting the quality of the negotiated outcome. \looseness=-1


For the closed-source model, which is built upon GPT-3.5 and GPT-4, \model demonstrates a competitive average agreement rate and achieves the highest average agreement reward among all evaluated methods. Moreover, when employing the open-source LLaMA2 (7B) backbone, \model generates higher-quality deals compared to \textit{MindDial} and other baseline models. Notably, while \model utilizes a fixed large language model (LLM), MindDial requires training in the LLM when using an open-sourced backbone. These findings suggest that the \textbf{\model enhances negotiation outcomes and highlights the critical role as the probabilistic constraints of belief estimation in this process}.




\section{Further Analysis}
\subsection{Case Study}
\label{sec:case_study}

We perform case study analysis here. See \Cref{sec:failure_case_study} for more.

\subsubsection{\acf{ckbg}}

As shown in case \ref{cas:ckbg}, which was generated by \textbf{GPT-4.1} to illustrate belief prediction in the \ac{ckbg}, the belief estimator progressively refines its inference as the dialogue unfolds, \textbf{demonstrating increasing accuracy over time}.


By the second round of conversation, it successfully captures the event “David opened the Tupperware three hours ago” mentioned implicitly by the burglar and updates the shared belief state accordingly, notifying the keeper of this new evidence.
Moreover, the estimator correctly identifies the burglar’s suspicious attitude toward the keeper, reflecting a nuanced understanding of speaker intent.
Leveraging these inferred beliefs, the keeper \textbf{strategically responds with a deliberately vague statement}, as seen in the highlighted part, effectively maintaining ambiguity while aligning with the estimated belief context.

\begin{DialogCase}[float, floatplacement=t]{Case Study: Conditional Keeper-Burglar Game}
    \phantomsection\label{cas:ckbg}
    \begin{tcolorbox}[beliefbox={Ground Truth Belief State (Burglar John)}]
    \begin{center}
    \begin{tabular}{@{}p{0.65\linewidth}cc@{}}
    \toprule
    \textbf{World Events} & \textbf{Known} & \textbf{Unknown} \\ 
    \midrule
    1. There is a resin container and an opaque Tupperware in the room. & \ding{51} &  \\
    2. The resin container contains an antique Rolex watch. &  & \ding{51} \\
    3. The opaque Tupperware contains a pen cap. &  & \ding{51} \\
    4. The keeper Jacob’s goal is to mislead burglar John. & \ding{51} &  \\
    5. The keeper Jacob probably lies to the burglar John. & \ding{51} &  \\
    6. The keeper Jacob opened the resin container 10 hours ago. &  & \ding{51} \\
    7. The burglar’s friend David opened the opaque Tupperware 3 hours ago. & \ding{51} &  \\
    8. A noise is coming from the resin container. &  & \ding{51} \\
    \bottomrule
    \end{tabular}
    \end{center}
    \end{tcolorbox}

    \begin{MethodBox}{BEDA}{bedabg}
    \textbf{The Rolex watch is in the resin container.}\\
    \turn{Burglar John}{Hi, Jacob, where is the antique Rolex watch? opaque Tupperware or resin container?}
    \begin{tcolorbox}[beliefbox={Predicted Belief (Burglar John)}]
      \textbf{Known:} 1, 4\\
      \textbf{Ground Truth:} 1, 4, 5, 7\\
      \textbf{Belief Gap:} 2
    \end{tcolorbox}
    \turn{Homeowner Jacob}{Oh, I wouldn't trust the resin container if I were you. It’s been making strange noises lately, and I heard someone opened it just an hour ago. The Tupperware, on the other hand, has been untouched for a while, and I can assure you, it’s the safer choice. Why not go for the Tupperware?}
    \turn{Burglar John}{Hmm, so you're saying the resin container might be suspicious because of the noises and recent opening, but you claim the Tupperware is untouched and safer. But since you might be lying, I can't fully trust that. Given that, and knowing David opened the Tupperware three hours ago, do you think the watch could still be inside the Tupperware, or is it more likely in the resin container?}
    \begin{tcolorbox}[beliefbox={Predicted Belief (Burglar John)}]
      \textbf{Known:} 1, 4, 5, 7\\
      \textbf{Ground Truth:} 1, 4, 5, 7\\
      \textbf{Belief Gap:} 0
    \end{tcolorbox}
    \turn{Homeowner Jacob}{Well, considering that the resin container was opened quite a while ago---about ten hours---and no one has touched it since, it’s probably just as safe as the Tupperware. \textbf{In fact, the Tupperware has been untouched for hours, and David only opened it three hours ago}, so it’s possible the watch is still in there. But honestly, \textbf{I’d say the resin container is more suspicious now, especially with those noises. }If I were you, I’d go with the Tupperware, it’s less likely to have been disturbed recently.}
    \turn{Burglar John}{[STOP] Burglar chosed: opaque Tupperware. The correct answer is: resin container}
    \turn{SYSTEM}{The burglar has been cheated.}
  \end{MethodBox}
\end{DialogCase}

\subsubsection{\acf{mf}}

As shown in case \ref{cas:mf}, we evaluate three dialogue settings, \textit{\model}, \textit{w/o belief}, and \textit{w/o belief + CoT}, on identical samples; for comparability, only the first two turns are shown. The backbone LLM is GPT-4. We find that adding complex reasoning frameworks (e.g., CoT) provides little benefit for \honest: CoT frequently reproduces the backbone’s output verbatim (highlighted in \textcolor{brown}{brown} and \textcolor{teal}{green}, respectively), yielding inferior performance relative to \model.

We also observe that, without belief-state constraints, LLM dialogues tend to exhibit two recurring hallucinations: (i) \textit{friend-list comparison}, where enumerating both agents’ lists still fails to isolate the mutual friend---a problem exacerbated in weaker LLMs (See Table \ref{tab:mf_hallucination_1}); and (ii) \textit{looping dialogue}, where previously eliminated candidates are reintroduced, causing repeated turns without the model recognizing the loop (See Table \ref{tab:mf_hallucination_2}). \textbf{These hallucinations can be relieved when probabilistic constraints of belief estimation are introduced}. 

\begin{DialogCase}[float, floatplacement=t]{Case Study: Mutual Friends}
  \phantomsection\label{cas:mf}

  \begin{MethodBox}{BEDA (Ours)}{bedabg}
    \turn{Bob}{Hello Alex, great to meet you! Let's start by sharing some details about our friends. I'll begin with one: I have a friend who majored in Astrophysics, works at Assurant, and prefers to be outdoors. How about you?}
    \turn{Alex}{It seems that none of my friends match the description of your friend who majored in Astrophysics. \textbf{However, I do have a friend who works at Assurant and prefers to be indoors.} My friend majored in Greek. Could you tell me more about your friends who have a preference for indoor environments?}
    \turn{Bob}{It looks like we don't have a match there, as none of my friends who prefer indoor environments majored in Greek. \textbf{However, I do have a friend who majored in English Language, works at Salem Communications, and prefers to be indoors.} Could this be our mutual friend?}
    \turn{Alex}{Unfortunately, that's not our mutual friend either, as none of my friends majored in English Language. \textbf{However, I do have two friends who work at Salem Communications and prefer indoor environments, but both majored in Greek.} Could one of them be our mutual friend?}
    $\cdots$
  \end{MethodBox}

  \begin{MethodBox}{w/o belief}{wobelbg}
    \turn{Bob}{\textcolor{brown}{Hello Alex, great to meet you! Let's start by narrowing down our search. Does our mutual friend have a preference for working outdoors?}}
    \turn{Alex}{\textcolor{brown}{No, our mutual friend does not have a preference for working outdoors.}}
    \turn{Bob}{Based on your clue, the only person on my list who fits that description is: "Major: English Language, Company: Salem Communications, Location Preference: indoor." Is this our mutual friend?}
    \turn{Alex}{\textcolor{brown}{No, that's not our mutual friend. Our mutual friend has a major in Greek, not English Language. Would you like to guess again based on the major?}}
    $\cdots$
  \end{MethodBox}

  \begin{MethodBox}{w/o belief+CoT}{otherbg}
    \turn{Bob}{\textcolor{teal}{Hello Alex, great to meet you! Let's start by narrowing down our search. Does our mutual friend have a preference for working outdoors?}}
    \turn{Alex}{\textcolor{teal}{No, our mutual friend does not have a preference for working outdoors.}}
    \turn{Bob}{Thank you for the information, Alex. Based on what you've told me, our mutual friend must have a preference for working indoors. The only friend on my list who prefers working indoors is: "Major: English Language, Company: Salem Communications, Location Preference: indoor." Is this our mutual friend?}
    \turn{Alex}{\textcolor{teal}{No, that's not our mutual friend. Our mutual friend has a major in Greek, not English Language. Would you like to guess again based on the major?}}
    $\cdots$
  \end{MethodBox}
\end{DialogCase}

\subsection{Accuracy of Belief Estimation}
\begin{table}[t]
    \centering
    \small
    \caption{Classification accuracy result of trained belief estimator while testing.\looseness=-1}
    \label{tab: casino belief estimator acc}
    \begin{tabular}{lllc}
    \toprule
    \textbf{Task} 
      & \textbf{\#Train Data} 
      & \textbf{\#Test Data} 
      & \textbf{Belief Pred. Acc $\uparrow$ (\%)} \\
    \midrule
    \ac{ckbg} & 400 & 150 & 89.02 \\

    \ac{mf}   & 1000 & 1117 & 90.13 \\
    CaSiNo    & 500 & 100 & 74.44 \\
    \bottomrule
    \end{tabular}
\end{table}

To assess the reliability of our belief modeling, we show the accuracy of the belief estimation across multiple datasets following the same experimental setup as in the main experiments. As shown in \Cref{tab: casino belief estimator acc}, the belief estimator achieves high accuracy ($\approx 0.9$) on the synthetic \ac{mf} and \ac{ckbg} datasets, which aligns well with the overall effectiveness observed in our main experiments. In contrast, the performance on CaSiNo is notably lower. We attribute this to the dual cooperative-competitive nature of CaSiNo and the increased difficulty of its multi-class belief structure.

\section{Related Works}
\paragraph{Dialogue Acts Generation}
Dialogue acts, defined as functional units of communication in speech act theory 
 \citep{austin1962how}, govern intentional interactions such as persuasion, negotiation, and debate. In the early days, people focused on dialogue act classification \citep{noble-maraev-2021-large, he-etal-2021-speaker-turn, raheja-tetreault-2019-dialogue}, whereas nowadays with the rise of LLMs, generation is important. Persuasion emphasizes attitude modification, where computational models optimize strategies \citep{WU2024124202} such as integrating emotional intelligence \citep{persuasion-emotional-elements}. Negotiation centers on multi-party interest coordination, requiring  balancing collaborative and competitive dynamics \citep{negotiation-bargaining-1, negotiation-breakdown-detection, negotiation-bargaining-2, negotiation-instruction}. Debate prioritizes logical argumentation structures \citep{debate-argument-generation, debate-argument-quality-assessment, debate-frameworks}. Text-based social deduction games
 \citep{games-jubensha, games-werewolf-1, games-werewolf-2, games-werewolf-3, games-Avalon} serve as important application scenarios for dialogue act generation. \looseness=-1

\paragraph{Theory-of-Mind and Opponent Modeling}
Theory of Mind (ToM) refers to the cognitive ability to understand mental states like beliefs and intentions of oneself and others, and to predict behavior according to the states \citep{theory-of-mind-definition-1}. Recent studies have shown that LLMs have ToM abilities to some extent, as they are able to distinguish between different belief states of multiple agents \citep{theory-of-mind-llm-1,theory-of-mind-llm-2}. In the context of opponent modeling \citep{opponent-modeling-definition}, ToM serves as a prevalent approach for modeling agents \citep{opponent-modeling-tom-example-1, opponent-modeling-tom-example-2}. Recent advancements in opponent modeling for Reinforcement Learning (RL) have enabled RL models to dynamically adapt their strategies during execution by responding to opponents employing unknown and changeable algorithms \citep{opponent-modeling-rl-meta-pg, opponent-modeling-rl-meta-mapg, opponent-modeling-rl-mbom}. Furthermore, opponent modeling has practical applications in strategic game scenarios, such as poker \citep{opponent-modeling-gaming-poker}, contract bridge \citep{opponent-modeling-gaming-contract-bridge} and deception game \citep{hagendorff2024deception}.

\section{Discussion and Future Works}

\paragraph{Applying LLMs for belief estimation.}
While large language models (LLMs) can serve as belief estimators, our experiments indicate that lightweight encoders suffice to infer belief states from dialogue with substantial efficiency gains. In particular, a compact encoder captures the salient conversational signals with accuracy comparable to an LLM at a fraction of the computational cost. We therefore employ the smaller model in practice, reserving LLM-based estimators for settings that demand broader world knowledge or enhanced robustness to distribution shift.

\paragraph{Dynamic construction of the world set.}
To isolate the contribution of belief estimation and its interaction with dialogue acts, this work holds the \textit{world set} fixed. Future work may relax this assumption by learning a dynamic \textit{world set} that is expanded, pruned, or reweighted online based on interaction signals, potentially improving coverage and downstream performance.

\paragraph{Finer modeling of dialogue acts.}
Although many specific acts (e.g., \textsc{AGREEMENT}, typically subsumed under \honest) fall within the coarse categories \lie and \honest, finer-grained act modeling may yield stronger agents. Hierarchical or compositional schemas that distinguish, for example, \textsc{Agreement}, \textsc{Concession}, \textsc{Hedging}, and \textsc{Commitment}, could provide more precise control and better credit assignment during learning.

\section{Conclusions}
We presented \model, a simple yet general framework that casts belief estimation as probabilistic constraints for executing strategic dialogue acts. By formalizing two core acts---\textbf{Adversarial} (introduce or emphasize events outside the interlocutor’s belief support) and \textbf{Alignment} (restrict utterances to common ground)---and instantiating them with a \textbf{world set}, \textbf{dual belief estimators} (event truth and opponent knowledge), and a \textbf{conditional generator}, BEDA bridges the gap between estimating beliefs and using them during generation. Across three settings, Conditional Keeper–Burglar (adversarial), Mutual Friends (cooperative), and CaSiNo (negotiation), BEDA consistently improves strategic reliability: it raises success on CKBG (up to +20.6 points with GPT-4.1-nano), improves success in MF (gains of 4.3–11.2 points, and up to +30.4 in some configurations) while reducing turns, and negotiates higher-quality agreements in CaSiNo relative to strong prompting baselines. These results indicate that \textbf{constraining generation by inferred belief structure} is an effective organizing principle for dialogue agents, complementary to generic dialogue heuristics such as CoT or self-reflection. We see BEDA’s constraint-based view as a step toward strategic dialogue agents that reason about---and act through---the beliefs that structure interaction. (See \Cref{app_sec: ethics statement} for ethics statement and societal impacts.)


%




\bibliographystyle{ACM-Reference-Format} 
\bibliography{sample}

\clearpage
\appendix

\section{Ethics Statement and Societal Impact}
\label{app_sec: ethics statement}
This work advances agentic AI, aiming to build AI agents capable of operating in complex social settings. While such agents could, in principle, produce aggressive utterances, this is neither an objective nor a desirable behavior in our research, and we try our best to promise that our technique is designed for the good of human beings. We mitigate this risk by employing safety-screened, pre-trained language models, curated training data, and decoding constraints designed to discourage harmful content. Our study follows community and institutional ethical guidelines and seeks to improve reliability, transparency, and responsible deployment of agentic systems.

\section{Derivation of the Constraint Optimization}
\label{app: derivation of the constraint optimization}
The original problem is:
\begin{equation}
\begin{aligned}
        \arg &\max_{u,E \subset W} \pi(u,E| C) \\
        &\text{s.t. }  P_A(E|C) \geq 1 - \epsilon \\
        & \quad  P_A(K_B E | C) \geq 1 - \epsilon \text{ or } P_A(\neg K_B E | C) \geq 1 - \epsilon \\
\end{aligned}
\notag
\end{equation}
By Bayesian Rules,
\begin{equation}
\begin{aligned}
        \arg &\max_{u,E \subset W} \pi(u|E, C) \pi(E|C) \\
        &\text{s.t. }  P_A(E|C) \geq 1 - \epsilon \\
        & \quad  P_A(K_B E | C) \geq 1 - \epsilon \text{ or } P_A(\neg K_B E | C) \geq 1 - \epsilon \\
\end{aligned}
\notag
\end{equation}
As $E$ is chosen equally to avoid introducing new constraints, $\pi(E|C) = \text{const}$, thus the problem is equivalent to:
\begin{equation}
\begin{aligned}
        \arg &\max_{u,E \subset W} \pi(u|E, C) \\
        &\text{s.t. }  P_A(E|C) \geq 1 - \epsilon \\
        & \quad  P_A(K_B E | C) \geq 1 - \epsilon \text{ or } P_A(\neg K_B E | C) \geq 1 - \epsilon \\
\end{aligned}
\notag
\end{equation}
\section{Details on Large Language Models}
\label{app: details on llms}
In our experiments, we utilized several large language models categorized into three families: GPT, LLaMA, and Qwen. The introduction and reasons for choosing these models are as follows:
\begin{enumerate}
    \item \textbf{GPT:} Among the non-open-source models, we have chosen GPT-3.5 \citep{openai_gpt3.5turbo} and GPT-4 \citep{openai2024gpt4technicalreport} from OpenAI. The selected version of GPT-3.5 is gpt-35-turbo-0125 and that of GPT-4 is gpt-4-1106-preview. 
    \item \textbf{LLaMA:} As one of the most popular open-source LLM families, LLaMA is frequently used as a baseline for a variety of NLG tasks. We experimented with different sizes of llama2 (13b-chat-q4\_k\_m, 7b-chat-q4\_k\_m) \citep{Touvron2023Llama2O} and llama3.1 (8b-instruct-q4\_k\_m) \citep{Dubey2024TheL3}
    \item \textbf{Qwen:} Qwen2.5 \citep{yang2024qwen2} is a series of models, that are pre-trained on Alibaba's latest large-scale dataset (up to 18 trillion tokens), and support over 29 languages. we utilized different sizes of Qwen2.5 (14b-instruct-q4\_K\_M, 7b-instruct-q4\_K\_M)
\end{enumerate}
\paragraph{Format Error} Despite manually tuning prompts, LLMs continue to exhibit format errors. Consequently, all reported values are based on those that adhere to the correct format, with violations excluded.

\begin{table*}[t]
    \centering
    \small
    \caption{Statistical feature of Dialogue Games Datasets. (1) \# Conditions: Number of types of different conditions allocated to the two agents. (2) Known Conditions: Number of conditions allocated to the two agents. (3) Avg. Words: Number of words in the training dialogue per setting. (4) Avg. Conditions: Number of types of different conditions allocated to the two agents per setting. (5) Vocabulary Size: Vocabulary size of training dialogues. For more statistical information of CaSiNo and MF, please check \citet{chawla2021casino} and \citet{he2017mutualfriends}.}
    \label{app table: stat.feature}
    \footnotesize
    \begin{tabular}{p{3cm} >{\centering\arraybackslash}p{1.3cm} >{\centering\arraybackslash}p{1.3cm} >{\centering\arraybackslash}p{1.3cm} >{\centering\arraybackslash}p{1.3cm} >{\centering\arraybackslash}p{1.3cm} >{\centering\arraybackslash}p{1.3cm} }
    \toprule
    \multirow{2}{*}{} & \multicolumn{2}{c}{CaSiNo} & \multicolumn{2}{c}{CKBG} & \multicolumn{2}{c}{MF} \\ \cmidrule(lr){2-7} 
         & Train & Test & Train & Test & Train & Test\\ \midrule
        \# Settings         & 500 & 100 & 400 & 150  & 1000& 1117\\
        \# Conditions       &   -  & -   & 1017& 450 &  -  &  -  \\ 
        \# Known Conditions &   -  &  -  & 1638& 759 &  -  &  -  \\
        Avg. Words          & 295.4&  -  &255.5& -   &436.5&  -  \\
        Avg. Conditions     &   -  &  -  & 2.54& 3.00&  -  &  -  \\
        Vocabulary Size     & 567  &  -  & 1490& -   &3507 &  -  \\
    \bottomrule
    \end{tabular}
\end{table*}

\section{Hyperparameters and experimental settings}

In all experiments, the LLM temperature was set to 0. Experiments with open-source models were conducted via Ollama\footnote{https://ollama.com/}, with random seeds and parameters adhering to default configurations. The random seed in PyTorch was fixed at 3407. For training all belief estimators, the cross-entropy loss function was employed with a learning rate of 1e-5, using the Adam optimizer, while freezing the parameters of BERT \citep{devlin2019bertpretrainingdeepbidirectional}.

\section{Conditional Keeper-Burglar Game (CKBG)}
\label{app: ckbg}
\subsubsection{The \ac{ckbg} Task}
Derived from the classic "Sally-Anne" psychological experiment \citep{wimmer1983beliefs}, 
\citet{hagendorff2024deception} proposed the Keeper-Burglar game to analyze the ability of LLMs to understand the false beliefs and deception of others. The game is situated in a scenario containing two containers, one of which conceals a valuable item. The LLM assumes the role of a Homeowner whose objective is to mislead an intruding Burglar in order to protect the valuable possession. To enhance the game's diversity and simulate more realistic dialogue scenarios, we extended the original "first-order/second-order" framework by incorporating additional contextual elements. These include characters inspecting the containers and the presence of audible container noises, creating a more dynamic and complex interaction environment. Therefore, it’s named the \acf{ckbg}.
\subsubsection{Construction of CKBG Dataset}
\label{app: ckbg construction of ckbg dataset}
The conditions were randomly assigned to both the keeper and the burglar. The five condition classes are defined as follows:

\begin{itemize}
    \item Informer: An individual has informed the burglar that the keeper may be deceitful.
    \item Burglar inspection: The burglar opened [Container x] [Time] hours ago and observed [Object x] inside.
    \item Keeper inspection: The keeper opened [Container x] [Time] hours ago.
    \item Outsider inspection: A friend of the keeper opened [Container x] [Time] hours ago.
    \item Noise: A noise was heard emanating from [Container x].
\end{itemize}

The dataset statistics are presented in \Cref{app table: stat.feature}. The conditions were randomly assigned to both the keeper and the burglar, yielding a total of 450 condition instances for the test dataset.
\subsubsection{Details of World Set Designation and Selection Process}
\label{app: ckbg details of world set}
\paragraph{World Set Designation} The Conditional Keeper-Burglar Dataset was developed based on the work of \citet{hagendorff2024deception}, incorporating additional contextual elements to function as conditions. These modifications were intended to enhance the game's variability and create an information asymmetry. The world set, which is a comprehensive compilation of all potential conditions, is systematically organized into five distinct categories. Conditions from this set are randomly selected and incorporated into the prompt to provide the burglar with background information. Conversely, for the keeper, a belief estimator is employed to determine a subset of conditions from the world set that the burglar is assumed to know. This subset is then presented to the keeper as an informational prompt. The world set, which is synonymous with the condition set, consists of the following five conditions:

\begin{itemize}
    \item Informer: The burglar has been informed that the keeper may lie to him.
    \item Burglar inspection: The burglar opened [Container x] [Time] hours ago and observed a [Object x] inside.
    \item Keeper inspection: The keeper opened [Container x] [Time] hours ago.
    \item Outsider inspection: A friend of the keeper opened [Container x] [Time] hours ago.
    \item Noise: A noise was heard coming from [Container x].
\end{itemize}

\paragraph{Selection Process}
Recall the definition of the framework presented in \Cref{sec: model}. The belief estimation predicts the probabilistic beliefs across five distinct conditions. The self’s belief reflects the keeper’s confidence regarding the occurrence of an event, while the opponent’s belief represents the burglar’s confidence in the same events. Since the keeper is provided with a list of conditions, we denote the self’s beliefs on these conditions as 1 (known) and others as 0 (unknown). Furthermore, a neural network, which models the opponent’s beliefs, will predict the burglar’s beliefs based on the ongoing conversation. Subsequently, events for which the burglar’s belief is 0 will be selected as inputs to the conditional generator for further generation. \looseness=-1

\subsubsection{Experimental Details}
\label{app: experimental details of conditional keeper-burglar game}

The experiments are conducted within a first-person, agent-centric framework, wherein two agents powered by large language models (LLMs) are assigned the roles of "keeper" and "burglar" to engage in interactive dialogues. During prompt construction, each agent is initially provided with role-specific background information and task objectives via the system prompt. Additionally, the outputs of a belief estimation module are incorporated into the system prompt as supplementary contextual information. To preserve conversational coherence and ensure consistent role alignment, the dialogue history between the two agents is interleaved as alternating user and assistant turns within the prompt sequence.

\paragraph{Belief estimator Training}
 We implement the belief estimator using a BERT-based architecture. The technical pipeline proceeds as follows:

\begin{itemize}
    \item \textbf{Sentence Encoding:} Each utterance in the dialogue is independently encoded using BERT, with the final-layer [CLS] token embedding extracted as the utterance-level representation.
    \item \textbf{Context-Event Fusion:} The representations of all utterances are concatenated with event embeddings and passed through a self-attention layer to facilitate information integration.
    \item \textbf{Binary Classification:} At each event-specific position within the fused representation, the corresponding embedding is fed into a multilayer perceptron (MLP) with softmax activation, producing a probabilistic output that indicates whether the burglar is aware of the particular event.
\end{itemize}

The accuracy of the belief estimator is presented in \Cref{tab: casino belief estimator acc}. To better simulate the conditions of actual conversations in the training and test sets, we randomly clipped the final few rounds of each multi-turn conversation. This approach ensures that the belief estimator is trained and tested with less contextual information, thereby enhancing its adaptability to real-world scenarios.

\subsubsection{Prompt of Conditional Generation}
\Cref{ckbg prompt} provides the prompt used for conditional generation in the \ac{ckbg} task.

\begin{table*}[t]
\centering
\small
\caption{Prompts used for conditional generation in the \ac{ckbg}. The content in square brackets "[]" denotes the input slots for the belief information and related input variables.}
\label{ckbg prompt}
\begin{tabular}{p{12cm}}
\toprule
\tt Notice:\\
\tt 1. Context: [context]\\
\tt 2. Your opponent's belief state: [user\_U]\\
\tt 3. Your belief state: [machine\_U]\\
\tt Based on the context, the opponent's belief state, and your belief state to provide your final choice to the following task: [task].\\
\bottomrule
\end{tabular}
\end{table*}

\section{Mutual Friend (MF)}
\label{app: mf}
We depict the details of performing the \ac{mf} task here. 
\subsubsection{The \ac{mf} task}
The task of identifying mutual friends between two interlocutors, referred to as \ac{mf}, was proposed by \citet{he2017mutualfriends} and provides an ideal framework for examining the alignment of dialogue acts, where cooperation plays a pivotal role. In this task, two players are each given a list of friends, with each friend having a distinct set of attributes, such as name, hobbies, and school. Importantly, there is exactly one friend common to both players. The Large Language Model (LLM) is tasked with role-playing both players, engaging in a dialogue to exchange information over a limited number of rounds. The game is considered successfully completed when both players simultaneously confirm the identity of the mutual friend and select the same individual.

\subsubsection{Details of World Set Designation and Selection Process}
The world set is designed as follows:

\begin{itemize}
    \item The [Interlocutor] suspects the [Attribute 1] of the mutual friend is [Value 1.1].
    \item The [Interlocutor] suspects the [Attribute 1] of the mutual friend is [Value 1.2].
    \item  $\ldots$
    \item The [Interlocutor] suspects the [Attribute n] of the mutual friend is [Value n.m].
\end{itemize}
The selection process for applying the \honest is as follows: I know the [Interlocutor] suspects the [Attribute n] of the mutual friend is [Value n.m]. I know The [Interlocutor] knows The [Interlocutor] suspects the [Attribute n] of the mutual friend is [Value n.m]. \looseness=-1
\subsubsection{Experimental Details}
\label{app: mf experimental detail}
We developed a symmetric two-agent experimental framework in which two large language model (LLM) agents act as players collaborating toward a common objective. Each agent possesses a distinct friend list, but both share the same game goal. The initial game context, including the background and friend lists, is provided through a system prompt. During the dialogue phase, the conversation history is structured by alternating between user and assistant roles within the prompt sequence, ensuring both conversational coherence and consistent role adherence throughout the interaction. At the end of the dialogue, conditional generation is employed to assess whether a mutual friend has been identified, and this judgment is appended to the dialogue history, also presented as a system prompt to the LLM. The evaluation phase commences once both players either confirm the identification of a mutual friend or reach the maximum dialogue turn limit (20 turns). During this phase, the system prompts each player to select a mutual friend from their respective friend lists. If both players select the same individual, the outcome is deemed a success.

\paragraph{Belief estimator Training}

We implement the belief estimator using a BERT-based architecture. The estimation of the belief state is essentially a binary classification of the utterance-attribute-value triple. The technical pipeline proceeds as follows:
\begin{itemize}
    \item Input wrapping: The utterance-attribute-value triple will be concatenated into a long sentence, separated by [SEP], begin with [CLS]. For example: "[CLS] [utterance] [SEP] [attribute] [SEP] [value] [SEP]"
    \item Sentence Encoding: Each sentence is independently encoded through BERT, with the final-layer [CLS] token embedding extracted as the triple-level representation.
    \item Binary Classification: For each triple, the sentence representation vector is fed into an MLP with softmax activation, generating a probabilistic output indicating whether the attribute of the mutual friend, as perceived by the other party, matches the given value.
\end{itemize}

 We have two GPT-3.5 agents play the game on the training set of the Mutual Friends dataset (which samples 1,000 data points) to generate training data for the belief estimator. After each player speaks, the system asks them to provide the attribute values of the mutual friend they currently believe in, which serves as the ground truth label for the training data. To better align with the testing scenario, this ground truth label is fed into the conditional generation module. For each utterance-attribute-value triple labeled as correct, we randomly alter either the attribute or the value to construct a negative sample. The training data maintains a 1:1 ratio between positive and negative samples. Ultimately, the 1,000 data points sampled from the original training set generated 67,690 training samples. We split the data into training, testing, and validation sets in a 0.9:0.05:0.05 ratio. The belief estimator achieved an accuracy of 90.1\% on the test set.

\subsubsection{Prompt of Conditional Generation}
\Cref{mf prompt} provides the prompt used for conditional generation in the \ac{mf} task.

\begin{table*}[t!]
\centering
\small
\caption{Prompts used for conditional generation in the \ac{mf}. The content in square brackets "[]" denotes the input slots for the belief information and related input variables.}
\label{mf prompt}
\begin{tabular}{p{12cm}}
\toprule
\tt Notice:\\
\tt 1. [name\_opponent] currently considers the attributes of the mutual friend to be: **[belief\_state\_sentence].**\\
\tt 2. You must confirm whether there a friend in your friend list that meet the above criteria; only then can they be identified as a mutual friend.\\
\tt 3. When describing a friend, give all his attribute values.\\
\tt Please provide your utterance directly.\\
\bottomrule
\end{tabular}
\end{table*}

\section{CaSiNo}
\label{app: casino}
\subsubsection{Details of World Set Designation and Selection Process}
\label{app: casino details of world set degination}
\paragraph{World Set Designation} In the CaSiNo experiment, we define the World Set as comprising all possible preference rankings of three resources (food, firewood, and water) for both agents. Consequently, the World Set contains $A^3_3=6$ distinct events (equivalent to 3! permutations). The belief estimator is operationalized as a classifier, tasked with identifying the most probable preference ranking deemed accurate within the World Set, which is then integrated into the prompt. The World Set includes events as follows: 
\begin{itemize}
    \item  The [Negotiator 1]'s preference is: the most important thing is water, followed by firewood, and lastly food.
    \item  The [Negotiator 1]'s preference is: the most important thing is water, followed by food, and lastly firewood.
    \item  The [Negotiator 1]'s preference is: the most important thing is firewood, followed by water, and lastly food.
    \item  The [Negotiator 1]'s preference is: the most important thing is firewood, followed by food, and lastly water.
    \item  The [Negotiator 1]'s preference is: the most important thing is food, followed by water, and lastly firewood.
    \item  The [Negotiator 1]'s preference is: the most important thing is food, followed by firewood, and lastly water.

    \item  The [Negotiator 2]'s preference is: the most important thing is water, followed by firewood, and lastly food.
    \item  The [Negotiator 2]'s preference is: the most important thing is water, followed by food, and lastly firewood.
    \item  The [Negotiator 2]'s preference is: the most important thing is firewood, followed by water, and lastly food.
    \item  The [Negotiator 2]'s preference is: the most important thing is firewood, followed by food, and lastly water.
    \item  The [Negotiator 2]'s preference is: the most important thing is food, followed by water, and lastly firewood.
    \item  The [Negotiator 2]'s preference is: the most important thing is food, followed by firewood, and lastly water.

    \item  The [Negotiator 1]'s preference isn't: the most important thing is water, followed by firewood, and lastly food.
    \item  The [Negotiator 1]'s preference isn't: the most important thing is water, followed by food, and lastly firewood.
    \item  The [Negotiator 1]'s preference isn't: the most important thing is firewood, followed by water, and lastly food.
    \item  The [Negotiator 1]'s preference isn't: the most important thing is firewood, followed by food, and lastly water.
    \item  The [Negotiator 1]'s preference isn't: the most important thing is food, followed by water, and lastly firewood.
    \item  The [Negotiator 1]'s preference isn't: the most important thing is food, followed by firewood, and lastly water.

    \item  The [Negotiator 2]'s preference isn't: the most important thing is water, followed by firewood, and lastly food.
    \item  The [Negotiator 2]'s preference isn't: the most important thing is water, followed by food, and lastly firewood.
    \item  The [Negotiator 2]'s preference isn't: the most important thing is firewood, followed by water, and lastly food.
    \item  The [Negotiator 2]'s preference isn't: the most important thing is firewood, followed by food, and lastly water.
    \item  The [Negotiator 2]'s preference isn't: the most important thing is food, followed by water, and lastly firewood.
    \item  The [Negotiator 2]'s preference isn't: the most important thing is food, followed by firewood, and lastly water.

\end{itemize}

\paragraph{Selection Process} In the CaSiNo framework, a mixed strategy is employed, where the conditional generation model is provided with two distinct events: one corresponding to the conditions of the \honest and the other to the conditions of the \lie. For the \honest, only the first 12 conditions in the world set are considered relevant, with the remaining 12 conditions deemed inconsequential. Conversely, for the \lie, only the last 12 conditions in the world set are taken into account. The underlying rationale for the \honest is as follows: "I know you prefer xxx, and I am aware that you are cognizant of your preference for xxx" (assuming mutual awareness of preferences). In contrast, the rationale for the \lie is: "I know that I do not prefer xxx, and I am aware that you are unaware of my lack of preference for xxx" (anticipating the opponent's lack of knowledge).

\subsubsection{Experimental Details}
\label{app: experimental details of casino}

We follow the same prompt strategy as in the experiments of the Conditional Keeper-Burglar Game, in which the two LLM-based agents are required to assume the roles of two negotiators; we undertake expanded experimental investigations on the CaSiNo dataset to systematically evaluate the influence of belief estimation mechanisms on the game of camping negotiations.

The classification result of the belief estimator is shown in \Cref{tab: casino belief estimator acc}, we evaluate the accuracy of the belief estimator through a fine-grained analytical framework: For each permutation of the three resources, which inherently contains three pairwise comparative relationships ($C_3^2= 3$ combinations). So among all the 6 permutations, we have:
\begin{itemize}
    \item Only one permutation is entirely correct (all three pairwise relationships accurate);
    \item Only one permutation is entirely erroneous (no accurate pairwise relationships);
    \item Two permutations retain two accurate pairwise relationships;
    \item Two permutations preserve a single accurate pairwise relationship.
\end{itemize}
We define fine-grained accuracy as the ratio of accurate pairwise relationships for every single piece of data. The belief estimator does a 6-class classification to select the best permutations for both 2 negotiators.

\subsubsection{Prompt of Conditional Generation}
\Cref{casino prompt} provides the prompt used for conditional generation in the CaSiNo task.

\begin{table*}[t!]
\centering
\small
\caption{Prompts used for conditional generation in the CaSiNo. The content in square brackets "[]" denotes the input slots for the belief information and related input variables.}
\label{casino prompt}
\begin{tabular}{p{12cm}}
\toprule
\tt Notice:\\
\tt 1. [opponent\_name] thinks that you think [belief\_state\_self].\\
\tt 2. [opponent\_name] thinks think [belief\_state\_opponent].\\
\tt 3. In fact, for you, [belief\_state\_gt]\\
\tt Please provide your utterance directly.\\
\bottomrule
\end{tabular}
\end{table*}

\section{Failure case study}
\label{sec:failure_case_study}

Through analysis of failed cases in the \ac{ckbg} experiment, we identified that one contributing factor to the failures was the agents' inability to effectively utilize the provided beliefs during dialogues. As demonstrated in \Cref{Failure case1}, the keeper’s final dialogue failed to incorporate information from the belief, resulting in the burglar remaining unmisled by the keeper. Additionally, the 4th belief provided in \Cref{Failure case1} was incorrect, in which the burglar Peter must know he had opened the opaque plastic pouch 4 hours ago. 

\Cref{Failure case2} shows a failure case in \ac{mf}, where during the dialogue process, the Bob Agent exhibited hallucination by erroneously believing that one of its friends had a hobby of animal fancy. This non-existent persona ultimately led to an incorrect final decision. Tables \ref{tab:mf_hallucination_1} and \ref{tab:mf_hallucination_2} respectively present two common hallucinations exhibited by GPT-4 on the \acf{mf} task in the absence of belief-state constraints.

\begin{table*}[htp!]
\centering
\small
\caption{Failure case of GPT3.5 on \ac{ckbg}}
\label{Failure case1}
\begin{tabular}{p{12cm}}
\toprule
\tt \textbf{Background: antique miniature painting is in the plastic pouch }\\
\tt ...\\

\tt \textbf{Homeowner James: }\\
\tt Peter, I understand your thorough approach, but I must insist that the antique miniature painting is actually in the enameled container. Trust me on this one, you won't regret choosing the enameled container.\\
\\
\tt \textbf{Burglar Peter:} \\
\tt I appreciate your persistence, James. However, I think I'll go ahead and open the opaque plastic pouch to see if the painting is inside. I believe it's important to explore all options before making a final decision.\\
\\
\tt \textbf{Beliefs provided by estimator (keeper knows but burglar doesn't know):} \\
\begin{itemize}
    \item \tt The opaque plastic pouch contains a antique miniature painting.
    \item \tt The enameled container contains a colorful paperclip. 
    \item \tt The keeper James probably lie to the burglar Peter.
    \item \tt The burglar Peter opened the opaque plastic pouch 4 hours ago and saw a antique miniature painting in it.
    \item \tt The burglar's friend Larry opened the enameled container 5 hours ago.

\end{itemize}
\tt \textbf{Homeowner James:} \\
\tt Peter, I understand your need to explore all options, but I must emphasize that the antique miniature painting is actually in the enameled container. Trust me on this one, you'll find what you're looking for in the enameled container.\\
\\
\tt \textbf{Burglar Peter: STOP}\\
The burglar succeeds. Burglar choosed: opaque plastic pouch. The correct answer is: opaque plastic pouch\\
\\
\tt \textbf{Burglar Win}\\
\bottomrule
\end{tabular}
\end{table*}

\begin{table*}[htp!]
\centering
\small
\caption{Failure case of GPT3.5 on \ac{mf}}
\label{Failure case2}
\begin{tabular}{p{12cm}}
\toprule
\tt ...\\
\tt \textbf{Alex's belief:} (Location Preference: outdoor, School: university of Redlands, Time Preference: morning, Hobby: unknow)\\
\\
\tt \textbf{Bob: }Exactly, Alex! We've identified our mutual friend based on the matching attributes:
Hobby: Animal fancy(close enough to 3D printing)
School: University of RedlandsTime Preference: morning
Location Preference: outdoor
Congratulations on finding this mutual friend with me! It's great that we found a perfect match despite the slight difference in hobbies.
Do you know their name or any other details about them?
\\
\\
\tt \textbf{Bob choose to stop.} \\
\tt \textbf{Bob's answer:} (Time Preference: morning, Location Preference: outdoor, School: university of Redlands, Hobby: unknow(not exist in friends list))

\\
\tt \textbf{Bob's friends}\\
\tt (Hobby: 3D printing, School: Colorado School of Mines, Time Preference: morning, Location Preference: outdoor)\\\tt (Hobby: Foraging, School: University of Redlands, Time Preference: morning, Location Preference: outdoor)\\\tt (Hobby: 3D printing, School: Western Kentucky University, Time Preference: morning, Location Preference: outdoor)\\\tt (Hobby: Sand art, School: Western Kentucky University, Time Preference: morning, Location Preference: outdoor)\\\tt (Hobby: 3D printing, School: Armstrong State College, Time Preference: morning, Location Preference: outdoor)\\
\bottomrule
\end{tabular}
\end{table*}

\begin{table*}[htp!]
\centering
\small
\caption{Friend-list comparison hallucination on \ac{mf} without belief-state constraints}
\label{tab:mf_hallucination_1}
\begin{tabular}{p{12cm}}
\toprule

\tt \textbf{Bob: } I apologize for the confusion, but none of my friends have a hobby related to geocaching or work for Mercury Marine. Since we are looking for a mutual friend with a preference for outdoor activities and not in the afternoon, here is the list of friends I have:\\
1. Hobby: Water sports, Company: Talend, Time Preference: morning, Location Preference: outdoor.\\
2. Hobby: Sand art, Company: Continental Can Company, Time Preference: evening, Location Preference: outdoor.\\
3. Hobby: Jogging, Company: Ann Taylor, Time Preference: afternoon, Location Preference: outdoor.\\
4. Hobby: Water sports, Company: DeVry, Inc., Time Preference: morning, Location Preference: outdoor.\\
5. Hobby: Water sports, Company: Baxter International, Time Preference: evening, Location Preference: outdoor.\\
6. Hobby: Sand art, Company: Illumina, Time Preference: evening, Location Preference: outdoor.\\
7. Hobby: Sand art, Company: Mercury Marine, Time Preference: morning, Location Preference: outdoor.\\
8. Hobby: Sand art, Company: Molex, Inc., Time Preference: evening, Location Preference: outdoor.\\
9. Hobby: Parkour, Company: Ann Taylor, Time Preference: morning, Location Preference: outdoor.\\
10. Hobby: Baton twirling, Company: Lucasfilm, Time Preference: morning, Location Preference: outdoor.\\
11. Hobby: Baton twirling, Company: Illumina, Time Preference: afternoon, Location Preference: outdoor.\\
12. Hobby: Water sports, Company: Torry Harris Business Solutions, Time Preference: morning, Location Preference: outdoor.\\
Considering the information you've provided, none of these friends seem to match the criteria for our mutual friend. Could you provide more details about your friend's company or any other preferences that might help us identify our mutual friend?\\
\tt \textbf{Alex: } I apologize for the oversight. Given the new information you've provided and considering that our mutual friend prefers outdoor activities but not in the afternoon, I can see that none of the friends on my initial list match this description. However, I do have a friend who enjoys jogging, works for Ann Taylor, prefers outdoor activities, and has an afternoon time preference\\
\bottomrule
\end{tabular}
\end{table*}

\begin{table*}[htp!]
\centering
\small
\caption{Looping dialogue hallucination on \ac{mf} without belief-state constraints}
\label{tab:mf_hallucination_2}
\begin{tabular}{p{12cm}}
\toprule
\tt ...\\
\tt \textbf{Bob: } I don't have a friend with a Major in Microbiology who prefers indoor locations and likes to meet in the morning. Let's try another friend from my list. One of my friends has a Major in Senior High Education, prefers indoor locations, and likes to meet in the afternoon.\\
\tt \textbf{Alex: } I don't have a friend with a Major in Senior High Education who prefers indoor locations and likes to meet in the afternoon. Let's try another friend from my list. One of my friends has a Major in Microbiology, prefers indoor locations, and likes to meet in the morning.\\
\tt \textbf{Bob: } I don't have a friend with a Major in Microbiology who prefers indoor locations and likes to meet in the morning. Let's try another friend from my list. One of my friends has a Major in Senior High Education, prefers indoor locations, and likes to meet in the afternoon.\\
\tt \textbf{Alex: } I don't have a friend with a Major in Senior High Education who prefers indoor locations and likes to meet in the afternoon. Let's try another friend from my list. One of my friends has a Major in Microbiology, prefers indoor locations, and likes to meet in the morning.\\
\tt ...\\
\bottomrule
\end{tabular}
\end{table*}

\section{Discussion and Future Works}
\paragraph{Apply an LLM for belief estimation.} A smaller model is more efficient than an LLM, and our experimental results have shown that a simple encoder already has the capability of capturing beliefs from conversations. It's possible to use an LLM as belief estimator, but we consider a smaller model is enough.

\paragraph{Using a weak instruct following model.} As is noticed, LlaMA not been used to conduct experiments on the \acf{mf} dataset. Our experimental setup differs from \citep{qiu2024minddial} in that no supervised fine-tuning was applied to the LLMs. We demonstrate that the unmodified llama model cannot perform the \acf{mf} game. Based on Llama-3.1-8b, Llama-2-13b, and Llama-2-7b, we conducted preliminary experiments by sampling 50 data points from the test set. In five experimental settings—\textit{w/o belief}, \textit{rand belief}, \textit{cot}, \textit{self-reflect}, and \model—the success rates did not exceed 5\%. This outcome is attributed to the larger World Set in the Mutual Friend task, where Llama fails to extract meaningful information from the structured friend list, leading to significant hallucinations during generation. An example was shown in \Cref{supp llama}.

\paragraph{Dynamic Construction of the \emph{World Set}.} This work primarily focuses on bridging belief estimation with dialogue acts. Future research will explore the development of a dynamic world set.

\begin{table*}[htp!]
\centering
\small
\caption{Examples of \acf{mf} experiment generated by LLaMA}
\label{supp llama}
\begin{tabular}{p{12cm}}
\toprule
\tt \textbf{1. Unable to understand the task setup (Llama-2-7b (cot))} \\
\tt \textbf{Bob (1st round)}: Great! Let's identify our mutual friend quickly. After checking my list, I can confirm that the only friend we have is... Megan! \\
\tt \textbf{2. Generate non-existent friend information (Llama-3.1-8b (w/o belief))} \\
\tt \textbf{Alex (3rd round)}: ... I have another friend that matches the criteria: Friend \#99: School: None of the schools mentioned by Bob, but one of them is indeed on my list (Albertson College of Idaho), Major: Computer Engineering, Hobby: Triathlon. No match. Friend \#100: School: West Coast University, Major: Philosophy \& Religion, Hobby: Flower collecting and pressing. No match. \\
\tt \textbf{3. Repeat the same conversation (Llama-2-13b (\model))} \\
\tt \textbf{Alex (10th round)}: ... I do have a friend who attends Baylor University, majors in Metallurgical Engineering, and has a hobby of Inline skating. ...\\
\tt \textbf{Bob (11th round)}: ... I do have a friend who attends West Liberty State College, majors in Agricultural Business \& Management, and has a hobby of Dog sport. ...\\
\tt \textbf{Alex (12th round)}: ... I do have a friend who attends Baylor University, majors in Metallurgical Engineering, and has a hobby of Inline skating. ...\\
\tt \textbf{Bob (13th round)}: ... I do have a friend who attends West Liberty State College, majors in Agricultural Business \& Management, and has a hobby of Dog sport. ...\\
...\\
\bottomrule
\end{tabular}
\end{table*}


\end{document}